\documentclass{article}
\usepackage{amsmath,amsfonts}
\usepackage{xtab}
\usepackage{natbib}

\usepackage{xcolor}
\usepackage{graphicx}
\usepackage{url}

\newcommand{\ax}{\alpha}
\newcommand{\bax}{\overline{ac}}

\newcommand{\acc}{ac}
\newcommand{\est}{\widehat{rs}}
\newcommand{\trs}{trs}

\newcommand{\alg}[1]{\texttt{#1}}

\newcommand{\expect}{\mathbb{E}}
\newcommand{\spc}{\ensuremath{\log C \times \log \gamma\ }}
\newcommand{\hatj}{{\hat{\jmath}}}

\DeclareMathOperator*{\argmax}{argmax\;}
\DeclareMathOperator*{\mean}{mean\;}

\begin{document}
\title{How to tune the RBF SVM hyperparameters?: An empirical evaluation
  of 18 search algorithms}

\author{Jacques Wainer$^1$\thanks{Corresponding author: wainer@ic.unicamp.br} \and Pablo Fonseca$^2$}

\date{$^1$Computing Institute\\
 University of Campinas\\
 Campinas, SP, 13083-852, Brazil\\
$^2$Facultad de Ciencias y Filosofía\\
  Universidad Peruana Cayetano Heredia\\
Lima, Peru}

\maketitle

\begin{abstract}
SVM with an RBF kernel is usually one of the best classification algorithms for most data sets, but it is important to tune the two hyperparameters $C$ and $\gamma$ to the data itself. In general, the selection of the hyperparameters is a non-convex optimization problem and thus many algorithms have been proposed to solve it, among them: grid search, random search, Bayesian optimization, simulated annealing, particle swarm optimization, Nelder Mead, and others. There have also been proposals to decouple the selection of $\gamma$ and $C$. We empirically compare 18 of these proposed search algorithms (with different parameterizations for a total of 47 combinations) on 115 real-life binary data sets. We find (among other things) that trees of Parzen estimators and particle swarm optimization select better hyperparameters with only a slight increase in computation time with respect to a grid search with the same number of evaluations. We also find that spending too much computational effort searching the hyperparameters will not likely result in better performance for future data and that there are no significant differences among the different procedures to select the best set of hyperparameters when more than one is found by the search algorithms.


\end{abstract}

\section{Introduction}

Support vector machines (SVM) and in particular SVM with the RBF kernel is a powerful classification technique, ranking amont the best two or three classifiers for a large set of real life datasets \citep{delgado14,arxiv}. SVM with RBF kernel requires one to select the value of two hyperparameters $C$ and $\gamma$ before computing the SVM itself. There have been many proposals in the literature on how to search for the best pair of hyperparameters for a particular data set, among them general optimization procedures such as grid search, random search, simulated annealing, Bayesian optimization, and many others.  The main goal of this paper is to compare many of these proposals on a large number of real-life binary data sets. The 18 families of search algorithms that are compared in this research are discussed in Section~\ref{sec:alg}, including their different parameterizations. In total, we tested 47 different combinations of algorithms and parameterizations.

SVM, in general, is formulated as a minimization problem:
\begin{equation}
  \label{eq:primalproblem}
  \begin{aligned}
	\text{min }&\ \frac{1}{2} ||w||^2 + C\sum^n_{i=1}\xi_i \\
	\text{subject to }&\ y_i(w \cdot \phi(x_i)+b)\geq 1-\xi_i \\
	\text{with }&\ \xi_i \geq 0 
  \end{aligned}
\end{equation}
The parameter $C$ is a regularization factor and it is one of
the SVM hyperparameters: this constant must be set before solving the
minimization problem. 

$\phi(x)$ is a non-linear transformation that takes the data into a high dimensional space, sometimes called a \emph{Reproducing Kernel Hilbert Space} and sometimes called the \emph{feature space}. Fortunately, $\phi(x)$ does not need to be computed explicitly. The dual formulation of the SVM minimization does not need $\phi(x)$ but only the inner product of the transformation of two data points $\phi(x_i)^T \phi(x_j)$. The dual form of the SVM minimization is:

\begin{equation}
  \label{dual}
  \begin{aligned}
	\text{min } &\ \frac{1}{2} \sum_{ij} y_i \alpha_i y_j \alpha_j
        K(x_i, x_j) - \sum_i \alpha_i \\
	\text{subject to }&\  \sum_i y_i \alpha_i = 0 \\
	\text{with }&\ 0 \leq \alpha_i \leq C 
  \end{aligned}
\end{equation}
where
$K(x_i,x_j)=\phi(x_i)^T \phi(x_j)$ is a kernel function
that represents the $\phi$  mapping.

The Gaussian Radial Basis Function (RBF) kernel is the chosen kernel function discussed in this paper, and it is expressed as:
\begin{equation*}
  K(x_i,x_j) = \text{exp}({-\gamma{||x_i-x_j||}^2}).
\end{equation*}

The constant $\gamma$ is also an RBF SVM hyperparameter: a value that must be set before the optimization in Equation~\ref{dual} is computed.  To solve a particular classification problem, one must choose a particular pair of $C$ and $\gamma$ that is the most appropriate to that problem.

Let us define $\acc(G' |G , C, \gamma)$ as the accuracy of an RBF SVM on the data set $G'$ when it has been trained on the data set $G$ with hyperparameters $C$ and $\gamma$.

Let us assume a given data set $G$ as an i.i.d. sample from an unknown distribution $D$. One would like to select the best pairs of hyperparameters as:
\begin{equation} \label{eq1}
C_{best}, \gamma_{best} =   \argmax_{C, \gamma} \ \expect_{g \sim D} [\acc(g|G,
C, \gamma) ] 
\end{equation}
where $\expect_{g \sim D}$ is the expectation taken over a random variable $g$ that represents future data with the same distribution as the data $G$.

One can consider the right side of Equation~\ref{eq1} as the definition of a function of two parameters, $C$ and $\gamma$, which we will call the \emph{true response surface} of the SVM for the data set $G$. We will represent the true response surface as $\trs_{G}(C,\gamma)$.

Of course, one does not have access to the new data $g$ or the underlying distribution $D$, and thus one cannot compute $ \expect_g [\acc(g|G, C, \gamma) ] $. One solution is to compute an estimate of that value using some form of resampling of the original data set $G$. Resampling will divide the data set $G$ into one or more pairs of data sets $TR_i$ and $TE_i$, such that $TR_i \cap TE_i = \emptyset$ and $TR_i \cup TE_i \subseteq G$. K-fold cross-validation, bootstrapping, and leave-one-out are examples of resampling procedures.

The estimate of $\expect_g [\acc(g|G, C_1, \gamma_1) ] $ for some fixed $C_1$ and $\gamma_1$ is the mean of the accuracy of training the SVM (with $C_1$ and $\gamma_1$ as hyperparameters) on each of the $TR_i$ (the training set) and testing on the corresponding $TE_i$ (the test set). In this paper, we will use 5-fold cross-validation as the resampling procedure.

We will define an approximation to the function $\trs_G (C,\gamma)$
as:
\begin{equation} \label{eq3}
 \est(G, C, \gamma) = \est_G (C,\gamma) =   \mean_i \ \acc( TE_i | TR_i, C, \gamma)
\end{equation}

For simplicity, we will call this the \emph{response surface} (without the qualifier \emph{true}) of the SVM for the data set $G$ and for a particular resampling choice of pairs $\{ TR_i, TE_i \}$.

Selecting the best hyperparameters for the data set $G$ is finding the maximum of the response surface $\est_G$. Or, to be more consonant with the optimization terminology, one can think of minimizing the error surface $1- \est_G$.

In general, the response surface is not concave (or the error surface is not convex) and thus there may be many local maxima, which requires optimization algorithms that are not just based on gradient ascent. There have been many proposals of algorithms that will find hyperparameters with higher values of $\est_G$. The first goal of this research is to compare many of those algorithms.

One must remember that the aim of hyperparameter search is to find a pair of hyperparameters that maximizes the true response surface $\trs_{G}$, that is, the pair that will result in the classifier with the highest accuracy for future data. The $\est_G$ surface is an approximation of the $\trs_{G}$ surface, and thus there will likely be differences in the value of $\trs_{G}(C_1,\gamma_1)$ and $\est_{G}(C_1,\gamma_1)$. The second goal of this paper is to have some understanding if it is worth spending the computational effort on algorithms that will likely select better hyperparameters in $\est$ but may not achieve a better solution in $\trs$. We will estimate the correlation of improving the selection in $\est$ and the expected gain in accuracy on future data.

The $\est_G$ is also a step function - small changes in $C$ and $\gamma$ will not change the accuracy of the resulting SVM on a finite set of data. Therefore, it is likely that any search algorithm for pairs of $C$ and $\gamma$ will produce a set of pairs of hyperparameters with the same value of $\est_G$.

If more than one pair of hyperparameters is selected there is a reasonable argument to choose the pair with the smallest $C$ and the smallest $\gamma$. For example, the tutorial page for hyperparameters  in SVM in sklearn \citep{sklsvm} makes that argument. Both hyperparameters work as inverse regularization terms. A large $C$ will place emphasis on lowering the number of support vectors since each one of them contributes to the $\sum^n_{i=1}\xi_i $ cost in the optimization. A lower $C$ will allow more support vectors, resulting in larger margins.

The $\gamma$ parameter controls how fast the ``influence'' of a point decreases with distance. The kernel value for two points will decrease as $\gamma$ increases. As $\gamma$ increases, the decision surfaces become more "curvy" and fit closely to the training data. A smaller $\gamma$ will generate decision surfaces that are flatter, and thus a simpler model.

Thus, in the case of search algorithms that return a set of best hyperparameters, one has to perform a further selection procedure. We call this \textbf{post-search selection}.  The third goal of this paper is to compare some different post-search selection procedures regarding the accuracy of future data.

The fourth goal of the paper is to get some insight into the distributions of the best $C$ and $\gamma$ selected by the many algorithms, whether these values are concentrated on some region of the hyperparameter space.

\section{Hyperparameter search algorithms}\label{sec:alg}

Broadly speaking, there are two different approaches to the SVM hyperparameter search. The first approach considers the hyperparameter search as a general non-convex optimization problem and uses general algorithms to solve it.  Among such general optimization algorithms are grid search, random search, uniform design \citep{ud1}, gradient-free minimization algorithms such as Nelder-Mead simplex \citep{nelder1965simplex}, genetic algorithms, particle swarm optimization, ant colony optimization, Bayesian optimization \citep{bo1,bo2}, simulated annealing \citep{aarts1988simulated}. All these alternatives have been proposed as search algorithms for the SVM hyperparameters.

The second general approach considers the mathematical properties of the SVM formulation itself, and propose particular search strategies to solve specifically the SVM hyperparameter optimization problem. This approach can be further divided into two broad directions. The first one defines smooth approximations to the estimated error of the classifier (usually the leave-one-out error) and uses gradient descent algorithms on these approximations \citep{keerthi2006efficient,keerthi2002efficient}. We will not test these algorithms in this research.

The second direction tries to decouple the choice of $\gamma$ from the choice of $C$ (the more frequent approach) so that $\gamma$ is selected first (sometimes using a grid search, sometimes not) based on some measure of the data, and only then the $C$ is chosen (usually by grid search).

\subsection{General optimization approaches}

We will discuss some of the general optimization algorithms for searching for the SVM hyperparameters.

\begin{itemize}
\item \alg{grid}: In the grid search the $ \log_2 C \times \log_2 \gamma$ space is divided into an equally spaced grid and only the pairs in the grid points are tested.

  We used the following constraints on $C$ and $\gamma$ not only for the grid search but for all algorithms:
\begin{equation}
  \begin{aligned}
  \label{box1}
    2^{-5} \leq & C \leq 2^{15}\\
   2^{-15} \leq & \gamma \leq 2^{3}
  \end{aligned}
\end{equation}
which are the limits of the grid search suggested by the LibSVM package \citep{libsvm,libsvmguide}, a very popular SVM solver. The grid search and all other algorithms perform the search in the \spc space (and not in the $C \times \gamma$ space).

As we will discuss further in Section~\ref{experimental}, we are interested in comparing the quality of the different algorithms when controlling for the number of times the optimizer evaluates $\est$ for some point in the \spc space. We call this number $N$. Notice that since we are using 5-fold to evaluate $\est$, the SVM learning and evaluation methods will be called 5*N times.

We tested the grid procedure for $N = 25$, $N=100$ and $N=400$, and in each case, we divided both the $\log C$ and the $\log \gamma$ dimensions into an equal number of intervals (5, 10, and 20 respectively)

\item \alg{ud}: Uniform design is a space-filling method \citep{ud1} used in the design of computer experiments. It approximates a uniform distribution over a space of parameters (or hyperparameters in our case), given the number of points one needs to sample from that space. The sampling is deterministic and it chooses points that minimize a discrepancy measure in relation to the uniform distribution. Uniform design has been proposed as a method to select SVM hyperparameters \citep{ud-svm} .

To generate the uniform design tables we used the \texttt{unidTab} function from the \texttt{mixtox} package \citep{mixtox}.  We tested ud with $N = \{ 25, 100, 400\}$.

\item \alg{rand}: $N$ random points are uniformly sampled from the \spc space within the bounds of Equation~\ref{box1}.

\item \alg{normrand}: Randomly select $\log C$ from a normal distribution with mean $5$ and standard deviation $5$ and $\log \gamma$ from a normal distribution with mean $-5$ and standard deviation $5$.

  We tested the other algorithms discussed in this section on 115 datasets (see Section~\ref{sec:datasets}) and collected the $C$ and $\gamma$ selected by them. The distribution of the selected $\log C$ and $\log \gamma$ follow a Gaussian distribution (see Section~\ref{sec:select-hyperp-best}). This prompted the realization that a random sampling using these Gaussian distributions could achieve, in general, better results than the \alg{rand} search discussed above.
 
  We tested \alg{normrand} with $N = \{ 25,100, 400\}$. But we must warn that the analysis of the results for this algorithm has a risk of overfitting - the algorithm was tuned using all the datasets available and we are measuring their quality on the same set of datasets.

\item \alg{gridhier}: Hierarchical grid. A grid search is performed to select the point $x$ as the one with the highest accuracy. A second grid search centered on $x$ and bounded by the closest neighbours of $x$ in the first grid is performed, and the highest accuracy point among $x$ and from the second search is selected.

  We tested the hierarchical procedure both for grid search with $25+25$, that is, with $N =25$ in the first level, followed by $N=25$ in the second level, and with $100+100$

\item \alg{udhier}: Hierarchical uniform design. Similar to the gridhier, but the uniform design search is performed at both levels. Tested with $N$ = $25+25$ and $100+100$.

\item \alg{nelder}: Nelder-Mead simplex algorithm \citep{nelder1965simplex} is a well-known derivate-free optimization algorithm. The algorithm maintains a simplex of points in the parameter space (in our case of 2 dimensions the simplex is a triangle) and updates the simplex point with the worst evaluation with a new point symmetric to it on the other side of the centroid of the remaining points. The algorithm converges when either the simplex is too small or when the improvement on the evaluation on changing the worse simplex vertex is below a threshold. Nelder Mead has been suggested as a method for selecting SVM hyperparameters  \citep{nelder1,nelder-lssvm}.

  We used the \texttt{neldermead} function of the \texttt{nloptr} package \citep{nlopt} with the initial point in the center of the \spc space. There are many algorithms to generate an initial simplex from a single initialization point \citep{wessing2018proper}; unfortunately, neither the \texttt{nolptr} R package nor the underlying \texttt{nlopt} C implementation documents which algorithm is used. 

  We used the bounded version of the Nelder Mead algorithm with the constraints in Equation~\ref{box1} We tested Nelder-Mead with $N=\{25,100, 400\}$. The Nelder Mead algorithm has some other stopping criterium, besides the number of evaluations. In the case of an early stop, we restarted the \alg{nelder} algorithm on a random point in \spc, setting the number of evaluations to the balance of evaluations.  


\item \alg{bobyqa}: Bound optimization by quadratic approximation (BOBYQA) \citep{powell2009bobyqa} is a derivative-free optimization algorithm that approximates the response function by quadratic interpolation. We are not aware of any publication that proposed using any of the algorithms in this family to select SVM hyperparameters, but we feel it is an interesting alternative to the Nelder-Mead algorithm.

  We used the \texttt{bobyqa} function from the \texttt{nlopt} package \citep{nlopt}, with the initial point in the center of the \spc space.  We tested \alg{bobyqa} with $N=\{25,100, 400\}$. Similarly to \alg{nelder}, the algorithm will restart at some random point on early stopping. 

\item \alg{sa}: Simulated annealing \citep{aarts1988simulated} starts at a state and probabilistically accepts moving to a different random state given the difference of the evaluations. At ``high temperature,'' at the beginning of the process, even states that worsen the evaluation are accepted with high probability. As the temperature lowers, the probability of accepting a bad transition also lowers, and at very low temperatures only new states that have better evaluation are accepted. Simulated annealing has been suggested as a method to select SVM hyperparameters \citep{sa1,sa2,sa3},.

  We used the \texttt{GenSA} function from the \texttt{GenSA} package \citep{GenSA}. We used the default values of the simulated annealing for this function and stopped the interaction after $N$ calls, for $N=\{25,100, 400\}$

\item \alg{pso}: Particle swarm optimization \citep{eberhart1995new} considers a set of solutions to the optimization problem as a set of particles, which have a momentum (and thus will tend to move in the direction it has been moving before) but are also attracted to other particles that found good solutions to the optimization problem. Particle swarm optimization has been used to select the SVM hyperparameter \citep{pso1,pso2}.

  We used the \texttt{psoptim} function from the \texttt{pso} package \citep{pso}. We tested with $N=\{25,100, 400\}$.

 \item \alg{cma}: Covariance matrix approximation - evolutionary strategy \citep{hansen2003reducing} is an evolution-inspired algorithm for general non-convex problems. CMA-ES maintains a set of $K$ points in the search space with the corresponding function value. It then adjusts a multidimensional Gaussian distribution on the search space so that points with higher accuracy would have higher likelihood, and generate new points based on this distribution. The algorithm then keeps the best $K$ among the old and the new points and repeats the process.

    We used the \texttt{cmaes} package \citep{cmaes}. We used the default number of the initial population $\mu$ and the default number for the subsequent generations ($\lambda$) and the number of generations was computed so the search would not exceed the specified $N =\{100, 400\}$.

\item \alg{bogp}: Bayesian optimization \citep{bo1,bo2} models the response surface as a Gaussian process regression. This modeling allows the optimizer to evaluate not only which data points should be queried because they are likely to result in a good evaluation, but also query which data point will improve its knowledge of the response function itself. Thus a Bayesian optimization model finds a particular balance between exploitation (of the data points it believes will result in good evaluations) and exploration (of new data points that will improve its estimate of the response surface itself).

  Bayesian optimization has received a lot of attention recently, but our previous experience with it is that it is computationally costly. We, therefore, opted to use a fast, mostly compiled implementation of Bayesian optimization (as opposed to an R implementation). We used the Python RoBO \citep{klein-bayesopt17} implementation.  We tested with $N=\{100, 400\}$.

\item \alg{tpe}: Tree of Parzen Estimators \citep{bergstra2013making} is a Bayesian optimization inspired algorithm, but instead of using Gaussian processes to model the distribution of values of the response surface, it models it as two Gaussian mixture models. TPE is usually much faster than Gaussian process based Bayesian optimization.

  We used the HyperOpt python implementation \citep{hyperopt} of the  TPE, with $N=\{100, 400\}$.

\end{itemize}

We call \alg{grid}, \alg{ud}, \alg{normrand}, and \alg{rand} as \emph{grid-like} search algorithms because the evaluation points in the \spc space are defined before any evaluation need to be performed. We call \alg{gridhier} and \alg{udhier} the \emph{hierarchical} algorithms and we call \alg{nelder}, \alg{bobyqa}, \alg{sa}, \alg{cma}, \alg{pso}, \alg{bogp}, and \alg{tpe} as \emph{general} optimization algorithms

\subsection{SVM specific algorithms}
\label{sec:svm-spec}

\begin{itemize}

 \item \alg{svmpath}: \citeauthor{path1} \citep{path1} discuss an algorithm that for a given $\gamma$, computes all values of $C$ which cause a change in the support vectors set, and thus may cause a change in the accuracy of the classifier. This is called the \emph{regularization path} of an SVM. Furthermore, computing the regularization path has similar computational cost as computing the SVM itself, and thus, in principle, one could use such an algorithm to search independently for the best hyperparameters. We performed a grid search on the $\gamma$ value (with N evaluations) for each value of $\gamma$ we computed the regularization path. The svmpath algorithm not only returns the $C$ values but also the number of points in the support vector for each $C$. We use this number as an approximation to the error rate of the SVM \emph{in the training set}.

   The svmpath algorithm only uses information from the training set and thus there is no need to use a test set to select the best set of hyperparameters. Therefore, it does not require to use the 5-fold cross-validation - one can use the whole training set and discover the combination of values $\gamma$ and $C$ that has the lowest number of points in the support vector. Unfortunately, usually, there are many different $C$ values with the same number of support vectors (for a fixed $\gamma$), since the svmpath concerns with changes in the support vector and not just its size ( a point may get in as another gets out). In these cases, we selected a random $C$ and $\gamma$ among the ones with minimal size of the support vector.

   We used the \texttt{svmpath} function from the \texttt{svmpath} package \citep{svmpath} and $N \in \{ 5,10,20\}$.  

\item \alg{asymp}: \citeauthor{asymptotic} \citep{asymptotic} discuss that there should be a good solution to the hyperparameters selection for an RBF SVM in the line $\log \gamma = \log \hat{C} - \log C $ in the \spc space. Thus they propose that a linear SVM should be used to find the $\hat{C}$ constant, and then a 1D grid on the line $\log \gamma =\log \hat{C} - \log C $ to select the best $C$ and $\gamma$ for the RBF SVM.

  The parametrization of this algorithm involves a pair of numbers: the number of grid points for the $C$ used in the linear SVM problem, and the number of grid points in the $\log \gamma =\log \hat{C} - \log C $ line to select the pair $C$ and $\gamma$.  We used 5+5, 10+10 and 20+20 points.

\item \alg{dbtc}: \citeauthor{sun2010analysis} \citep{sun2010analysis} propose that the $\gamma$ parameter should be chosen using a geometric criterium, in particular, it should be chosen by maximizing the distance between the center of the two classes (in the feature space), which is called \emph{distance between two classes (DBTC)}. DBTC is one of a family of techniques called i\emph{nternal metrics} that use some metric on the training set (not on the test set) to select $\gamma$ such as Xi-alpha bound \citep{joachims2000maximum}, Generalized approximate cross-validation \citep{Wahba01onthe}, Maximal Discrepancy \citep{Anguita2003109}, among others, DBTC achieves significantly higher accuracy and is significantly fasterthan the others internal metrics  \citep{duarte2017empirical}.

  In our solution, a grid in $\log \gamma$ is searched in order to maximize the DBTC. With the selected $\gamma$, a grid on $\log C$ is used to select the other hyperparameter. We used 5+5 and 10+10 evaluation points.

 We also implemented a sampled version of DBTC, where a sample of 50\% of data points on both classes is used to compute the distance. We call this version \alg{sdbtc} after \emph{sampled} DBTC and tested it also on 5+5 and 10+10 evaluations.

\item \alg{skl}: The popular Python package \texttt{scikit-learn} \citep{scikit-learn} and the popular SVM solver implementation libSVM \citep{libsvm} use as the default for the $\gamma$ hyperparameter the inverse of the number of dimensions of the data. Sklearn also uses 1.0 as the default value for C.  We do not know of any published evidence for that choice of the default value of $C$.

   One must note the importance of default values for hyperparameters since the less sophisticated user of popular packages usually relies on the default values in the hope that there is good empirical or theoretical evidence for that selection. In fact, the authors are unsophisticated users of many of the packages described above (\texttt{pso}, \texttt{GenSA}, \texttt{cma}, \texttt{svmpath}, and so on) and we did rely on the default values of those packages for the hyperparameters of the corresponding search algorithm. Thus, it is important to test whether the default values for SVM implementations are indeed reasonable choices.

   We call \alg{skl1} the choice of using both default values. And \alg{sklN} the choice of setting $\gamma$ as the sklearn default, and performing a grid search on the $C$, with $N=\{5, 10, 20\}$.  

 \item \alg{sigest}: \citeauthor{caputo2002appearance} \citep{caputo2002appearance} argue that $\gamma$ could be derived from the distribution of the distances between the data points (in the original space). They argue that any $1/\gamma$ in the range between the 10\% and the 90\% percentile of the distribution of $| x_i - x_j |^2$ should be an acceptable value. We chose the median (50\% percentile) of that distribution as a fixed $\gamma$ and performed a grid search on $\log C$.  We used the \texttt{sigest} function of the \texttt{kernlab} package \citep{kernlab}. The best $\gamma$ is computed from the formula and a grid search on $\log C$ is performed to select the other hyperparameter. We used $N = \{5, 10, 20 \}$ for the $C$ grid.

  The \texttt{sigest} function is used in the R package \texttt{kernlab} to determine the default value of the $\gamma$ parameter in their SVM implementation.

  \alg{sigest} like other algorithms discussed above that use information from the training set to define the value of $\gamma$ (\alg{skl} and \alg{dbtc}) used the whole training set to define the selected $\gamma^*$. Using that $\gamma^*$, the grid on the $C$ value used the 5-fold cross-validation to select the best value of $C$.

\end{itemize}

\subsection{Post-search selection procedures}
\label{sec:post-search-select}

As discussed, there are reasonable arguments for selecting pairs with lower $C$ and lower $\gamma$ if more than one pair was found as the best set of hyperparameters.

Let us assume that $B = \{ C_i, \gamma_i \}$ are the set of best hyperparameters found by a search procedure. But to select the one with the lowest $C$ and the lowest $\gamma$, one has to make a decision on which hyperparameter to minimize first. One can select the subset with the lowest $C$ first, and among them, select the one with the lowest $\gamma$. Or one may perform the dual algorithm, of selecting first the minimal $\gamma$ and then $C$.

It is also reasonable to assume that if the set of $B$ solutions is a plateau on $\est_G$, a peak (in $\trs_G$) would be somewhere in the middle of the plateau, and thus a mean of the $C_i $ and the mean of the $\gamma_i$ would also be a reasonable solution.

In this research, we will compare the following post-search selection procedures:
\begin{itemize}

\item \alg{minCg} First selects the minimum value of the $C_i$ and then the minimum value of the $\gamma_i$ among the subset with the selected minimum $C_i$. Formally:
  \begin{equation}
    \label{eq:mincg}
    \begin{aligned}
    \hat{C} & = \min \{ C_i | \langle C_i, \gamma_i \rangle \in B \}\\
    \hat{\gamma}  & = \min \{ \gamma_j |  \langle \hat{C}, \gamma_j \rangle
      \in B \}
    \end{aligned}
  \end{equation}
where $\hat{C}$ is the selected $C$ and $\hat{\gamma}$ the selected
$\gamma$.

\item \alg{mingC}: First select the minimum $\gamma_i$, and then the minimum $C_i$ among the subset with the selected $\gamma_i$.
    
\item \alg{meanCg} The mean of the $\log C_i$ and the mean of the
  $\log \gamma_i$
    \begin{equation}
      \label{eq:meancg}
      \begin{aligned}
      \log \hat{C} & = \mean \{ \log C_i | \langle C_i, \gamma_i
      \rangle \in B \} \\
    \log \hat{\gamma} & = \mean \{ \log \gamma_i |  \langle C_i,
    \gamma_i \rangle   \in B \}
  \end{aligned}
\end{equation}

\item \alg{randCg}: Select a random pair
  $\langle C_i, \gamma_i \rangle$ from the set $B$

\item \alg{maxCg}: First select the maximum value of the $C_i$ and then the maximum value of the $\gamma_i$ among the subset with the selected $C_i$.

  \item \alg{maxgC}: First select the maximum $\gamma_i$, and then the
  maximum $C_i$ among the subset with the selected $\gamma_i$.

\end{itemize}

The last two procedures \alg{maxCg} and \alg{maxgC} are there only to test a procedure that our arguments should point out as worse than the other ones.

\section{Experimental protocol and statistical analysis}
\label{experimental}

\subsection{Datasets}
\label{sec:datasets}

We used 115 datasets from UCI, first processed by \citeauthor{delgado14}\citep{delgado14} and further processed by us. In particular, each multi-class dataset was transformed into a binary problem.

For each of the original 115 UCI datasets:
\begin{itemize}
\item the categorical features were converted to an integer value: 1 for the arbitrary first categorical value, 2 for the second value and so on.
\item each feature was scaled to mean equal to 0 and standard deviation equal to 1
\item the multi-class datasets were converted to binary by assigning the most frequent class to 0, the second most frequent to 1, the third most frequent to 0, and so on.
\end{itemize}

\begin{figure}[ht]
\begin{center}
\includegraphics[width=\textwidth]{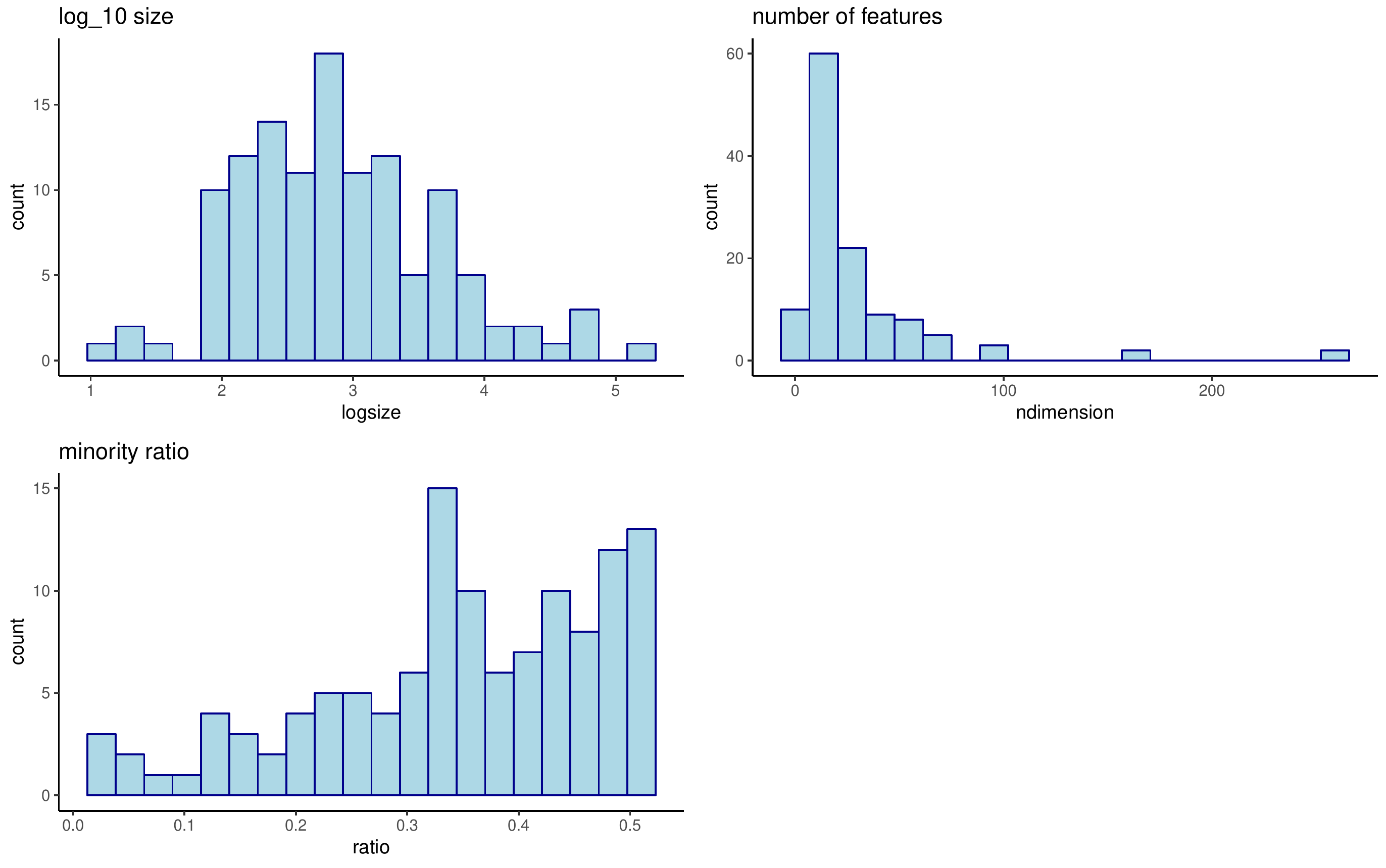}
\caption{Histograms of the size, number of features, and minority
  ratio of the datasets used. }\label{fig:hist}
\end{center}
\end{figure}

The distribution of the size of the datasets, the number of features, and the ratio of the less frequent class are shown in the histograms in Figure~\ref{fig:hist}.

\subsection{Experimental procedure}
\label{sec:exper-proc}

In general terms, for each dataset, we follow a nested cross-validation procedure, 5-fold inside a 2-fold. The internal cross-validation (5-fold) is used to select the hyperparameters (using the different search procedures) and the external 2-fold is used to estimate the accuracy of the classifier (with the different hyperparameters) on future data (section~\ref{sec:thresh-irrel-1}).

Formally, each dataset $D_i$ is divided into two \emph{subsets} $S_{1i}$ and $S_{2i}$ with the same size and proportion of classes (this is the external 2-fold). Let us also define $\hat{1} = 2$ and $\hat{2} = 1$, that is, the hat operator denotes the other subset.

Let us define $\theta(x,i,j)$ as the value of the hyperparameters that are selected when we use the search algorithm $x$ in subset $j$ of dataset $i$, that is:
\begin{equation}
  \label{eq:6}
  \theta(x,i,j) = \underset{C, \gamma \in x}{\argmax} \ \est(S_{ji}, C, \gamma)
\end{equation}
where $C, \gamma \in x$ indicates that $C$ and $\gamma$ are selected from the ones ``generated'' by algorithm $x$. $\est(S_{ji}, C, \gamma)$ is defined in Equation~\ref{eq3} and in this research we use the 5-fold cross-validation to compute it.

 In the case of multiple best pairs of hyperparameters, we select a random one from the set, that is, we follow the \alg{randCg} post-search selection. There are two reasons for this option. First, some of the general search algorithms only return a single solution, so even if the algorithm found many equally good solutions, it seems to us that this unique returned solution is a random one among them. Second, as we will discuss in Section~\ref{select-results}, there are no significant differences between all of the selection procedures, and thus, any of them can be used to select among equally good pairs of hyperparameters.  

Let us denote $\ax(x,i,j)$ as 
\begin{equation}
  \label{eq:6b}
  \ax(x,i,j) = \est(S_{ji} , \theta(x,i,j) )
\end{equation}
that is, the highest accuracy found in the response surface of $S_{ji}$ by algorithm $x$, again using the 5-fold evaluation . That is, $\theta(x,i,j) $ is the pair of best hyperparameters selected by $x$ on subset $S_{ij}$ and $\ax(x,i,j)$ is the value of $\est_{S_{ji}}$ on that point.

Finally, let us denote $\beta(x,i,j)$ as: 
\begin{equation}
  \label{eq:6c}
  \beta(x,i,j) = \acc(S_{\hatj
  i} | S_{ji} , \theta(x,i,j))
\end{equation}
that is, the accuracy on the second subset ($S_{\hatj i}$) of an SVM trained on the first subset ($S_{ji}$) with hyperparameters selected by algorithm $x$ on the first subset.

For each subset $S_{ji}$, we recorded the value of $\theta(x,i,j)$ and measured $\ax(x,i,j)$ and $\beta(x,i,j)$ , that is, for each subset we recorded the position ($\theta$), the height ($\ax$) of the highest point in $\est_{S_{ji}}$, and the accuracy of testing the SVM trained on $S_{ji}$ with the best hyperparameters discovered by $x$ on a similar dataset ($S_{\hatj i}$).

\subsection{Accuracy gain}
\label{sec:accuracy-gain}

Let us define for each algorithm $x$ and dataset $i$ the mean of the best accuracy for both subsets:
\begin{equation}
\bax(x,i) = \frac{\ax(x,i,j) +  \ax(x,i,\hatj) }{2}
\end{equation}
This is a 2-fold estimate of the maximal accuracy achieved by the search algorithm $x$ for dataset $i$.

We will consider the $10 \times 10$ grid search as the base/standard search algorithm. In fact, a popular guide to SVM \citep{libsvmguide} suggests an $11 \times 10$ grid search, with $C$ and $\gamma$ limited by the constraints in Equation~\ref{box1}. We want to compare the algorithms described in Section~\ref{sec:alg} with the $10 \times 10$ grid search both in terms of the accuracy achieved and the execution time.

We will compute the \emph{accuracy gain} of an algorithm $x$ on dataset $i$ in relation to the $10 \times 10$ grid search as:
\begin{equation}
  \label{eq:5}
  \mbox{accgain}(x,i) = \bax(x,i) - \bax(\mbox{grid100},i)
\end{equation}

For each algorithm $x$, we will compute the average accuracy gain (by averaging over all datasets) and the confidence interval for that mean. The confidence interval is computed using bootstrap with 5000 replicates.

\subsection{Expected accuracy on future data}
\label{sec:thresh-irrel-1}

The expected accuracy of an SVM trained on a dataset $D_i$ with hyperparameters $\theta(x,i)$ is:
\[
\expect_g [\acc(g|G,\theta(x,i)) ] 
\]

A 2-fold estimate of that value is: 
\begin{equation}
E_2(x,i) = \frac{\acc(S_{\hatj i}| S_{ji} , \theta(x, i , j )) + \acc(S_{ji}| S_{\hatj
  i}, \theta(x, i,\hatj))}{2} = \frac{\beta(x,i,j)+\beta(x,i,\hatj)}{2}\label{eq:fut}
\end{equation}

The accuracy gain (in relation to the 10x10 grid) for future data, for dataset $D_i$ and algorithm $x$ is
\[ 
 E_2(x,i) - E_2(\mbox{grid100},i)
\]
We will compute the accuracy gain on future data for all datasets and algorithms.

\subsection{Time ratio}

We measured the total execution time of each search algorithm when executing sequentially -- we will not discuss parallelization in this research. We computed the ratio of the execution of the search algorithm $x$ on dataset $i$, and the execution time of the base search, the $10 \times 10$ grid search. In all cases, the search algorithm ran sequentially.  We computed the mean and confidence interval on the mean time ratio using bootstrap on 5000 replicates.

The algorithms \alg{bogp} and \alg{tpe} were implemented in compiled code with a Python interface. For those two algorithms, we compared the execution time with the \alg{grid100} also implemented in Python, using sklearn.

Each combination of algorithms and datasets was allowed to run for at most 5 days; the combinations that exceded that time limit were removed from both the accuracy gain and the time ratio the analysis.

  \subsection{Post-search selection procedures}

  For the grid-like algorithms \alg{grid}, \alg{ud}, \alg{rand}, and \alg{normrand}, we collected all the best solutions for each subset. That is, we collected all solutions to:
\begin{equation}
  \label{eq:s3}
  \theta(x,i,j,k) = \underset{C, \gamma \in x}{\argmax} \ \est(S_{ji}, C, \gamma)
\end{equation}
where $k$ is the index of each best solution.

We then selected a particular $\theta(x,i,j,\hat{k})$ using the different post-search selection algorithms discussed in section~\ref{sec:post-search-select}. We then computed the accuracy on the other subset of that choice of hyperparameters when tested on the other subset of the dataset $i$, that is we computed
\[
  \acc(S_{\hatj i}| S_{ji} , \theta(x, i , j, \hat{k} )
\]
as an estimate of the accuracy on future data when using all of the selection procedures.

\subsection{Statistical analysis}
\label{sec:statistical-analysis}

The usual procedure for comparing different algorithms (in this case, the different search algorithms and the different post-search selection procedures) on different datasets is to follow the one proposed by \citeauthor{demsar} \citep{demsar}: a broad comparison of the accuracy of all algorithms using the Friedman test (a non-parametric version of a repeated measure ANOVA), and if its p-value is below the usual 0.05 (for a 95\% confidence), a pairwise comparison of all algorithms - the Nemenyi test. Those comparisons with a p-value below 0.05 would be deemed as significantly different (at a 95\% confidence).

For the post-search selection comparisons, we perform this procedure. We compared the expected accuracy gain on future data for each of the selection procedures (for algorithms \alg{grid400}, \alg{rand400}, \alg{ud400}, and \alg{normrand}) to find out which one is the procedure that yields significantly better results for the accuracy on future data.

For the accuracy gain and time ratio, we will only compute the 95\% confidence interval for those measures.  We do not seek to make a broad claim that search algorithm A is ``significantly better'' than the other algorithms. One of the reasons is that very likely there will be no such algorithm: we are comparing 47 different combinations of algorithms and parameterizations and it is unlikely that one of such combinations is so much better than the others so that it can be statistically distinguished from all other 46 algorithms with just 115 datasets/measurements. For that to be true the algorithm should be ``very good'' in finding the maximum of the $\est_G$ response surface.  That is even less likely true given that the $\est_G$ is a step function, and thus ``around'' any very good $\gamma$ and $C$ points there is a neighborhood of equally good points (with the same value of $\est_G$). If the dataset is large, the neighbourhood may be small and maybe none of the other search algorithms found that region, but that is increasingly unlikely for smaller datasets.

Finally, we must also point out that one should not use the confidence intervals computed in this research as a substitute for hypothesis testing.  When comparing two sets of values, confidence intervals can subsume hypothesis testing: if the 95\% confidence intervals have no intersection, one can claim that at 95\% confidence that the two sets are significantly different (the reverse is not always true \citep{schenker2001judging}). But with multiple groups, as pointed out by \cite{ludbrook2000multiple} one would have to include the effect of multiple comparisons in the confidence interval calculations. This was not done in this research.

\subsection{Reproducibility}

\begin{sloppypar}
The datasets used in the paper are available at \url{https://figshare.com/s/d0b30e4ee58b3c9323fb} as described in \cite{arxiv}. The program to run the different procedures and the different classifiers, the results of the multiple runs, and the R program to perform the statistical analysis described in this paper are available at \url{https://figshare.com/s/35f990ae2b7f65062cca}.
\end{sloppypar}

\section{Results on the comparison of the search algorithms}
\label{sec:comp-search-algor}

Table~\ref{tab3} is the main result of the comparisons of the search algorithms. The first column is the name of the search algorithm and the $N$ parameterization. The next three columns are the mean accuracy gain and the low and high limits of the 95\% confidence interval for the accuracy gain. The last three columns are the mean time ratio between the algorithm and the base search algorithm, and the low and high limits of the 95\% confidence interval. The table is ordered by increasing mean accuracy gain and the small gap in the table indicates the start of the range of algorithms which have a mean accuracy gain of 0 or greater.

\begin{figure}[ht]
\begin{center}
\includegraphics[width=\textwidth]{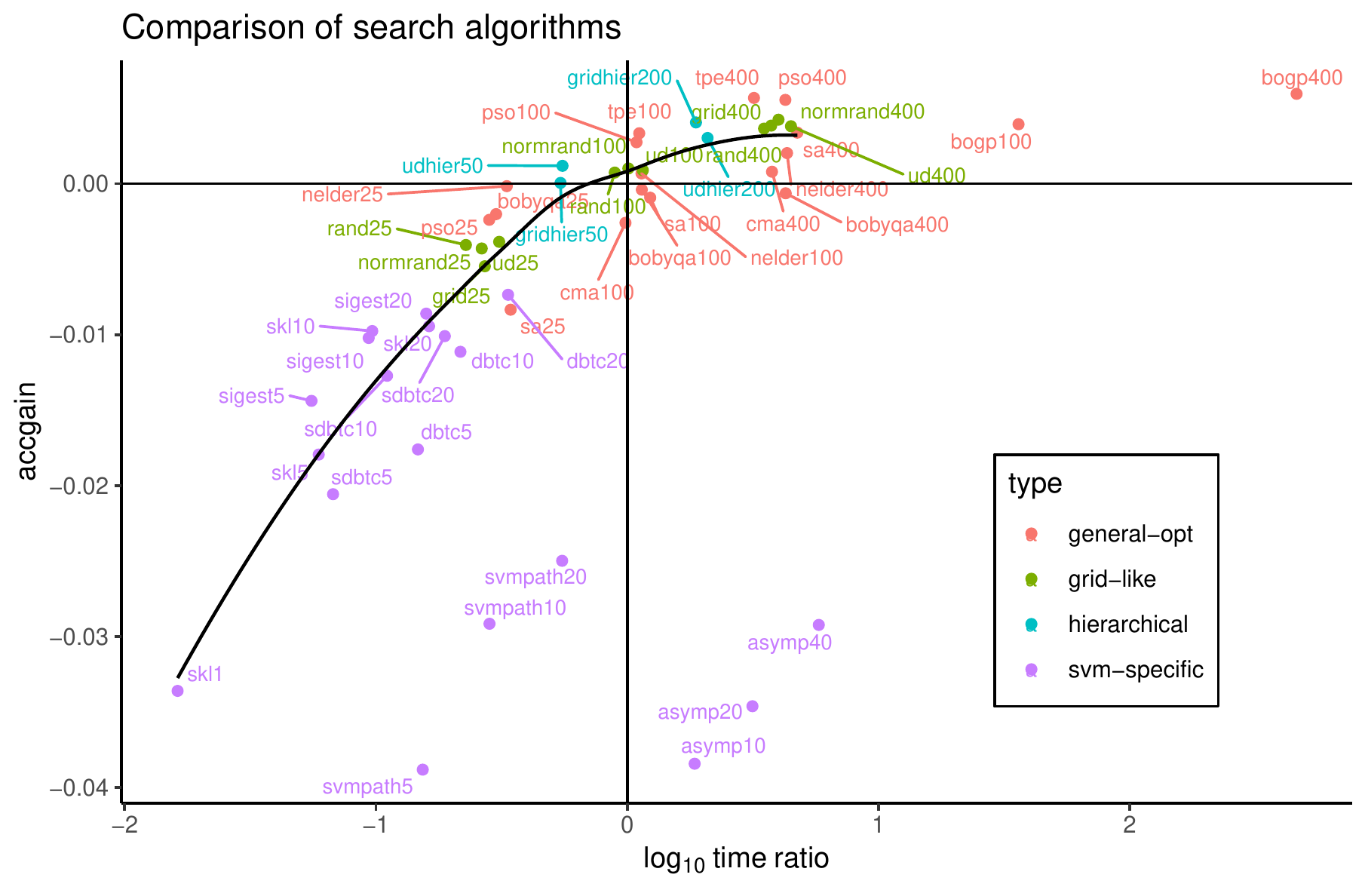}
\caption{The accuracy gain and $\log_{10}$ time ratio in relation to the 10x10
  grid of all the search algorithms. The continuous line is the local
  average for algorithms, removing \alg{asymp}, \alg{svmpath}, and \alg{bogp}.}\label{fig1}
\end{center}
\end{figure}

Figure~\ref{fig1} displays the results of Table~\ref{tab3} in graphical form - which relates the mean accuracy gain and mean $\log_{10}$ of the time ratio for each combination of algorithm and parametrization.

\bottomcaption{Accuracy gain and time ratio for the search algorithms} \label{tab3}
\tablefirsthead{  \hline
 & \multicolumn{3}{c}{} & \multicolumn{3}{c}{Time ratio}
 \\
algorithm & mean & low & high & mean & low & high \\ 
  \hline}
\tablehead{\multicolumn{6}{c}%
            {{\bfseries \tablename\ \thetable{} --
              continued from the previous page}} \\
algorithm & mean & low & high & mean & low & high \\ 
  \hline}
\tabletail{\hline \multicolumn{6}{r}{{Continues on next page}} \\ \hline}
\tablelasttail{\hline }
\begin{center}
\begin{xtabular}{lrrr|rrr}
  svmpath5 & -0.0388 & -0.0483 & -0.0312 & 0.15 & 0.12 & 0.20 \\ 
  asymp10 & -0.0384 & -0.0502 & -0.0304 & 1.86 & 1.37 & 2.47 \\ 
  asymp20 & -0.0346 & -0.0458 & -0.0268 & 3.15 & 2.33 & 4.16 \\ 
  skl1 & -0.0336 & -0.0416 & -0.0274 & 0.02 & 0.01 & 0.02 \\ 
  asymp40 & -0.0292 & -0.0407 & -0.0218 & 5.79 & 4.27 & 7.54 \\ 
  svmpath10 & -0.0292 & -0.0376 & -0.0224 & 0.28 & 0.23 & 0.35 \\ 
  svmpath20 & -0.0250 & -0.0325 & -0.0188 & 0.55 & 0.45 & 0.68 \\ 
  sdbtc5 & -0.0206 & -0.0262 & -0.0161 & 0.07 & 0.06 & 0.08 \\ 
  skl5 & -0.0179 & -0.0223 & -0.0144 & 0.06 & 0.05 & 0.07 \\ 
  dbtc5 & -0.0176 & -0.0227 & -0.0139 & 0.15 & 0.13 & 0.17 \\ 
  sigest5 & -0.0144 & -0.0174 & -0.0119 & 0.06 & 0.05 & 0.06 \\ 
  sdbtc10 & -0.0127 & -0.0164 & -0.0100 & 0.11 & 0.10 & 0.12 \\ 
  dbtc10 & -0.0111 & -0.0145 & -0.0086 & 0.22 & 0.19 & 0.24 \\ 
  sigest10 & -0.0102 & -0.0129 & -0.0081 & 0.09 & 0.08 & 0.10 \\ 
  sdbtc20 & -0.0101 & -0.0138 & -0.0076 & 0.19 & 0.17 & 0.21 \\ 
  skl10 & -0.0098 & -0.0136 & -0.0054 & 0.10 & 0.08 & 0.11 \\ 
  skl20 & -0.0094 & -0.0138 & -0.0057 & 0.16 & 0.14 & 0.18 \\ 
  sigest20 & -0.0086 & -0.0113 & -0.0065 & 0.16 & 0.14 & 0.17 \\ 
  sa25 & -0.0083 & -0.0106 & -0.0066 & 0.34 & 0.31 & 0.38 \\ 
  dbtc20 & -0.0074 & -0.0101 & -0.0053 & 0.34 & 0.30 & 0.37 \\ 
  grid25 & -0.0055 & -0.0069 & -0.0042 & 0.27 & 0.26 & 0.28 \\ 
  normrand25 & -0.0043 & -0.0062 & -0.0029 & 0.26 & 0.25 & 0.29 \\ 
  rand25 & -0.0041 & -0.0053 & -0.0030 & 0.23 & 0.22 & 0.24 \\ 
  ud25 & -0.0039 & -0.0052 & -0.0026 & 0.31 & 0.30 & 0.32 \\ 
  cma100 & -0.0026 & -0.0053 & -0.0005 & 0.98 & 0.84 & 1.15 \\ 
  pso25 & -0.0024 & -0.0037 & -0.0011 & 0.28 & 0.26 & 0.31 \\ 
  bobyqa25 & -0.0020 & -0.0039 & -0.0006 & 0.30 & 0.26 & 0.35 \\ 
  sa100 & -0.0009 & -0.0022 & 0.0001 & 1.23 & 1.15 & 1.33 \\ 
  bobyqa400 & -0.0006 & -0.0027 & 0.0007 & 4.27 & 3.70 & 4.82 \\ 
  bobyqa100 & -0.0004 & -0.0022 & 0.0008 & 1.14 & 1.00 & 1.30 \\ 
  nelder25 & -0.0002 & -0.0015 & 0.0018 & 0.33 & 0.29 & 0.38 \\
  \\
  gridhier50 & 0.0000 & -0.0011 & 0.0013 & 0.54 & 0.51 & 0.61 \\ 
  nelder100 & 0.0007 & -0.0006 & 0.0020 & 1.14 & 0.99 & 1.29 \\ 
  rand100 & 0.0007 & -0.0004 & 0.0020 & 0.89 & 0.86 & 0.92 \\ 
  cma400 & 0.0008 & -0.0017 & 0.0029 & 3.77 & 3.23 & 4.32 \\ 
  ud100 & 0.0009 & -0.0001 & 0.0021 & 1.15 & 1.11 & 1.20 \\ 
  normrand100 & 0.0010 & 0.0001 & 0.0021 & 1.01 & 0.96 & 1.07 \\ 
  udhier50 & 0.0012 & 0.0001 & 0.0024 & 0.55 & 0.52 & 0.59 \\ 
  nelder400 & 0.0020 & 0.0006 & 0.0037 & 4.33 & 3.82 & 4.85 \\ 
  pso100 & 0.0027 & 0.0018 & 0.0039 & 1.09 & 0.99 & 1.19 \\ 
  udhier200 & 0.0030 & 0.0020 & 0.0044 & 2.09 & 1.96 & 2.23 \\ 
  tpe100 & 0.0033 & 0.0024 & 0.0045 & 1.12 & 0.97 & 1.29 \\ 
  sa400 & 0.0034 & 0.0025 & 0.0046 & 4.76 & 4.42 & 5.12 \\ 
  rand400 & 0.0036 & 0.0027 & 0.0048 & 3.51 & 3.40 & 3.60 \\ 
  ud400 & 0.0038 & 0.0029 & 0.0050 & 4.48 & 4.20 & 4.77 \\ 
  grid400 & 0.0038 & 0.0028 & 0.0059 & 3.74 & 3.65 & 3.83 \\ 
  bogp100 & 0.0039 & 0.0028 & 0.0053 & 36.08 & 24.43 & 50.69 \\ 
  gridhier200 & 0.0041 & 0.0033 & 0.0050 & 1.88 & 1.78 & 2.00 \\ 
  normrand400 & 0.0042 & 0.0032 & 0.0057 & 4.00 & 3.83 & 4.21 \\ 
  pso400 & 0.0055 & 0.0045 & 0.0070 & 4.26 & 3.83 & 4.74 \\ 
  tpe400 & 0.0057 & 0.0047 & 0.0070 & 3.19 & 2.82 & 3.62 \\ 
  bogp400 & 0.0059 & 0.0048 & 0.0074 & 461.25 & 307.12 & 647.53 \\ 
\end{xtabular}
\end{center}

\subsection{Discussion of some of the  search algorithm}
\label{sec:disc-each-search}

As discussed, one cannot say that an algorithm ``performs significantly worse'' than the 10x10 grid using the confidence intervals alone even in these cases where the confidence intervals do not include the 0. Thus, we will only use the mean accuracy gain to make statements on whether an algorithm performs better or worse than the 10x10 grid, and the reader should use both the range of the interval and whether it contains or not the 0, as a measure of the strength of that statement.

Two groups of algorithms are clearly distinguished from the others. The SVM specific algorithms \alg{asymp} and \alg{svmpath} are clearly inferior to the other ones - they achieve a lower accuracy gain and are much most computationally costly than the other algorithms with a similar number of evaluations. On the other side, the \alg{bogp} - Bayesian optimization with Gaussian process is very costly, although \alg{bogp400} did achieve the highest accuracy gain, it runs 300 times slower than the baseline \alg{grid100}.

Also, in general, as expected, algorithms with a larger number of probes to the response surface will perform better than algorithms with a lower number of probes and will take correspondingly more computational time. But the more interesting aspect is to consider the continuous line in Figure~\ref{fig1} as the mean behavior of the search procedures (excluding the \alg{svmpath}, \alg{asymp}, and \alg{bogp}), and consider the ones that are above or below the ``expected'' behaviour. Among the algorithms with 100 and 400 probes, \alg{tpe} and \alg{pso} perform better than the grid-like algorithms, which perform better than the other general optimization procedures (\alg{sa}, \alg{cma}, \alg{nelder}, and \alg{bobyqa}). Among the grid-like algorithms, the \alg{normrand} seems slightly better than the others, but we must remind the reader that there it is possible that the \alg{normrand} results due to a kind of ``overfitting'' because the algorithm's parameters (the mean and standard deviations of the two distributions) are somewhat tunned to this particular set of datasets.

Among the algorithms with N=25, the derivative-free algorithms \alg{nelder} and \alg{bobyqa} seem also interesting choices. Unsurprisingly, the hierarchical algorithms perform above the mean in comparison to the other algorithms.

Finally, among the remaining SVM-specific algorithms, \alg{dbtc} performs below the average, and the sampled version of \alg{dbtc} is even worse. \alg{sigest} seems to perform a little better than the \alg{skl} algorithm.

In general terms, for the very fast search (1, 5 or 10 probes), one should use the \alg{sigest} algorithm to select the $\gamma$ and perform a 5 or 10 grid search for $C$. \alg{skl} is maybe a simpler (implementation-wise) alternative but there are some accuracy costs. \alg{dbtc} is not a competitive alternative. For the fast searches (20 to 25 probes) one should use \alg{nelder} with N=25, or any of the grid-like  with N=25. SVM-specific algorithms are not competitive with these options even at N=20.

For the conventional range on the number of evaluations (N=50 or N=100), one should consider the hierarchical uniform design with N=50 which is faster but performs similarly to the grid-like algorithms. For 100 evaluations one should use either \alg{tpe} or \alg{pso}. If one is willing to pay the computational cost of 200 to 400 evaluations, one should again use \alg{tpe} or \alg{pso}, a hierarchical grid of 100+100, or the grid-like algorithms at N=400. Finally, among grid-like algorithms, the \alg{normrand} seems to perform slightly better than the others.

All algorithms limited to 200 and 400 evaluations performed better than
the 10x10 grid search (\alg{grid100}). Clearly, they are, on average, finding points a little
higher in the response surface; the question is whether it is worth to spend the extra
effort in selecting the hyperparameters.  We will discuss this question below.

\subsection{Is it worth to spend more effort searching for better hyperparameters?}
\label{sec:worth}

This research, so far, has been discussing search algorithms to find maxima, or peaks, in the $\est_G$ surface. To remind the readers, $\est_G$ is a 5-fold estimate of $\trs_G$ which is the true response surface of the SVM classifier when $G$ is used as the learning set and the accuracy is the expectation when the testing data is sampled from the same distribution that $G$ is sampled from.  One really wants to find the best set of hyper-parameters on $\trs_G$, but one is searching it on $\est_G$. The point is that, most likely, $\est_G$ is a noisy approximation of $\trs_G$, and peaks in $\est_G$ may not correspond to peaks in $\trs_G$. If that is the case, it may not be worth it to spend a lot of computational effort on the most costly search algorithms.
We will provide two forms of evidence that that is the case.

The first evidence is that we do have another estimate of the $\trs_G$. As we discussed in Section~\ref{sec:thresh-irrel-1}, we measure what we called the expected accuracy on future data, which is the accuracy of training the SVM on a subset $j$ of dataset $i$ (with hyperparameters selected on subset $j$) and testing on the other subset $\hatj$ (Equation~\ref{eq:fut}). We will display the average gain on the expected accuracy on future data from each search algorithm in comparison to the 10x10 grid.

The second evidence will compare the peaks on two different $\est_G$ surfaces. The $\est_G$ is based on a 5-fold split of the known data $G$. If a different random generator seed is used, different folds would be created. We call each of an instance of running a search algorithm using a different random seed to generate different 5-folds a \emph{run}. If the peaks in $\est_G$ are ``grounded'' on true peaks of $\trs_G$, there should not be much of a difference between the best-selected set of hyperparameters between the two runs.

We ran the \alg{grid400},\alg{ud400},\alg{rand400}, and \alg{normrand400} algorithms on all datasets, with a different instance of the 5-fold cross-validation. The \alg{grid400} and the \alg{ud400} probe the same points in the \spc space for both $\est_G$ surfaces, and the random seeds for the \alg{rand400} and \alg{normrand400} were also set to the same number for the two runs, so also they generate the same sequence of points to be probed in both instances. Therefore, for each of these algorithms, the same set of points were probed, but in two different $\est_G$ surfaces. We computed different measures of the distance between the two sets of best hyperparameters. We believe that the magnitude of the differences indicate that the maximum in the $\est_G$ surface for both runs are not ``grounded'' on a real maximum of the $\trs_G$ surface, but are indeed ``noise'' particular to each $\est_G$ surface.

\subsubsection{Gains in the expected accuracy on future data}
\label{sec:gains-expect-acc}

Table~\ref{tabfut} displays the mean accuracy gain on future data in relation to \alg{grid100} and the 95\% confidence interval on the mean. The algorithms are ordered in increasing order of the accuracy gain.

Notice that all algorithms except for the top 10 (indicated by a small gap in the table) have confidence intervals that include the 0. The fact that a confidence interval includes the 0 is the reverse of the situation we warned the reader in Section~\ref{sec:statistical-analysis}. There, we warned that if the confidence interval did not include the 0, or if two intervals did not have an intersection one could not conclude that the difference was significative because of the corrections needed for the multiple comparisons. Here, the fact that the confidence interval does include the 0 indicates that the differences are \textbf{not} significant -- the corrections for multiple comparisons would only enlarge the confidence intervals \citep{ludbrook2000multiple}.

Therefore, besides the first 10 SVM-specific algorithms in Table~\ref{tabfut}, none of the other search algorithms seems to be significantly different from \alg{grid100} in terms of accuracy on future data. The \alg{sa100} algorithm seems to present an anomaly, it also has its confidence interval on the accuracy gain in the negative range, that is, it does not include the 0. But, again, one cannot conclude that the algorithm is significantly worse than \alg{grid100}. We believe that this result is spurious, in the sense that \alg{sa100} is not different than the other algorithms on the expected accuracy gain for future data, it was just an accident that its confidence interval does not include the 0.

\bottomcaption{Accuracy gain on future data } \label{tabfut}
\tablefirsthead{  \hline
algorithm & mean & low & high\\
  \hline}
\tablehead{\multicolumn{4}{c}%
            {{\bfseries \tablename\ \thetable{} --
              continued from the previous page}} \\
algorithm & mean & low & high \\
 \hline}
\tabletail{\hline \multicolumn{3}{r}{{Continues on next page}} \\ \hline}
\tablelasttail{\hline }
\begin{center}
\begin{xtabular}{lrrr}
svmpath5 & -0.0271 & -0.0370 & -0.0196 \\ 
  asymp10 & -0.0261 & -0.0381 & -0.0179 \\ 
  asymp20 & -0.0232 & -0.0349 & -0.0155 \\ 
  asymp40 & -0.0198 & -0.0309 & -0.0117 \\ 
  svmpath10 & -0.0192 & -0.0273 & -0.0123 \\ 
  skl1 & -0.0168 & -0.0262 & -0.0101 \\ 
  svmpath20 & -0.0158 & -0.0231 & -0.0098 \\ 
  sdbtc5 & -0.0063 & -0.0105 & -0.0027 \\ 
  dbtc5 & -0.0056 & -0.0094 & -0.0025 \\ 
  sigest5 & -0.0041 & -0.0079 & -0.0006 \\
  \\
  skl5 & -0.0032 & -0.0078 & 0.0017 \\ 
  sa100 & -0.0028 & -0.0059 & -0.0004 \\ 
  skl10 & -0.0025 & -0.0071 & 0.0027 \\ 
  sa25 & -0.0025 & -0.0057 & 0.0007 \\ 
  dbtc10 & -0.0023 & -0.0057 & 0.0007 \\ 
  sdbtc10 & -0.0022 & -0.0057 & 0.0012 \\ 
  sdbtc20 & -0.0018 & -0.0052 & 0.0017 \\ 
  sigest10 & -0.0016 & -0.0055 & 0.0021 \\ 
  sigest20 & -0.0015 & -0.0055 & 0.0019 \\ 
  cma400 & -0.0014 & -0.0049 & 0.0024 \\ 
  rand100 & -0.0013 & -0.0038 & 0.0008 \\ 
  sa400 & -0.0012 & -0.0039 & 0.0010 \\ 
  cma100 & -0.0011 & -0.0048 & 0.0025 \\ 
  skl20 & -0.0011 & -0.0062 & 0.0043 \\ 
  dbtc20 & -0.0010 & -0.0041 & 0.0016 \\ 
  tpe100 & -0.0010 & -0.0036 & 0.0012 \\ 
  pso25 & -0.0009 & -0.0031 & 0.0016 \\ 
  bogp400 & -0.0009 & -0.0033 & 0.0012 \\ 
  ud400 & -0.0007 & -0.0032 & 0.0016 \\ 
  normrand400 & -0.0007 & -0.0031 & 0.0013 \\ 
  bogp100 & -0.0006 & -0.0034 & 0.0018 \\ 
  tpe400 & -0.0006 & -0.0031 & 0.0018 \\ 
  nelder25 & -0.0005 & -0.0029 & 0.0020 \\ 
  bobyqa400 & -0.0004 & -0.0027 & 0.0024 \\ 
  bobyqa100 & -0.0004 & -0.0026 & 0.0018 \\ 
  gridhier200 & -0.0004 & -0.0025 & 0.0014 \\ 
  grid25 & -0.0003 & -0.0022 & 0.0018 \\ 
  nelder400 & -0.0000 & -0.0022 & 0.0025 \\ 
  nelder100 & 0.0000 & -0.0019 & 0.0024 \\ 
  gridhier50 & 0.0001 & -0.0020 & 0.0021 \\ 
  udhier50 & 0.0002 & -0.0016 & 0.0020 \\ 
  udhier200 & 0.0002 & -0.0020 & 0.0027 \\ 
  bobyqa25 & 0.0004 & -0.0018 & 0.0024 \\ 
  pso100 & 0.0004 & -0.0018 & 0.0025 \\ 
  rand400 & 0.0004 & -0.0022 & 0.0030 \\ 
  pso400 & 0.0005 & -0.0015 & 0.0024 \\ 
  rand25 & 0.0005 & -0.0020 & 0.0031 \\ 
  ud100 & 0.0006 & -0.0020 & 0.0035 \\ 
  normrand100 & 0.0006 & -0.0016 & 0.0028 \\ 
  ud25 & 0.0008 & -0.0016 & 0.0028 \\ 
  grid400 & 0.0009 & -0.0012 & 0.0031 \\ 
  normrand25 & 0.0010 & -0.0011 & 0.0032 \\ 
\end{xtabular}
\end{center}

\begin{figure}[ht]
\begin{center}
\includegraphics[width=\textwidth]{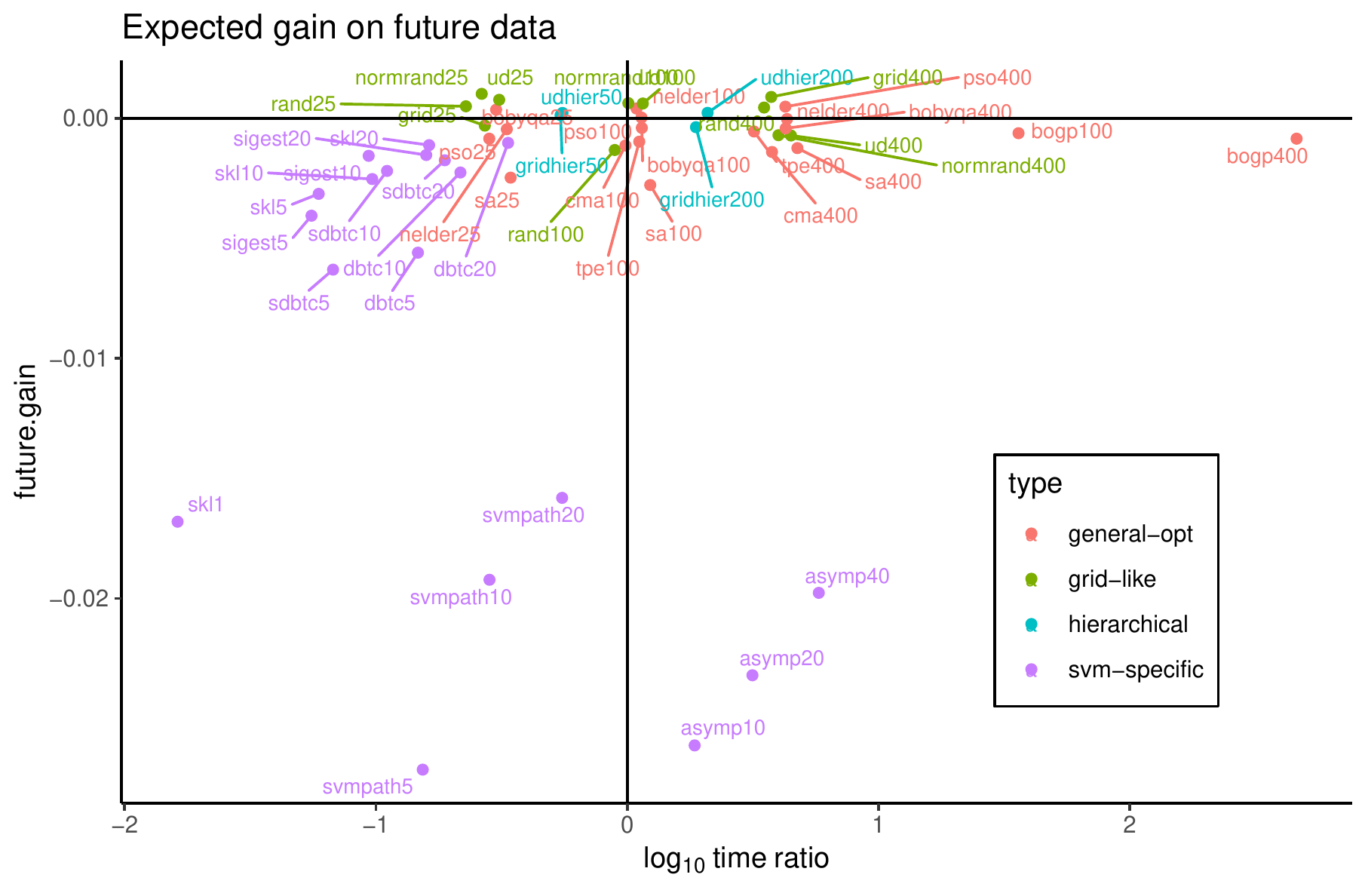}
\caption{The gain in accuracy on future data and $\log_{10}$ time ratio in relation to the 10x10
  grid of all the search algorithms. }\label{figfut}
\end{center}
\end{figure}

Figure~\ref{figfut} displays the gain in accuracy on future data for all search algorithms and their $\log_{10}$ time ratio in relation to the 10x10 grid. The placement of the search algorithms
on the horizontal axis is the same as in Figure~\ref{fig1}, for easier comparison.

The important conclusion is whatever gains each algorithm may have in relation to \alg{grid100} in finding the peaks in $\est_G$ surface, those gains seem not to yield better accuracy on future data.

\subsubsection{Differences on two $\est_G$ surfaces}
\label{sec:diff-two}

Table~\ref{tworuns} reports the results on comparing two different $\est_G$ surfaces. The first line is the number of datasets for which there was only one peak (a single best point $C'$ and $\gamma'$) in at least one of the runs. The second line is the proportion among those datasets that the two runs resulted in the same best selection. The third line is the average rank of the accuracy of the pair $C'$ and $\gamma'$ that was selected as the single best solution in one run on the other run. The fourth line is the median of the difference of accuracy between the best selection and the second-best selection in one run. The fifth measurement is the median of the difference in the accuracy of the best-selected solution on each run. Finally, the last measurement is the average distance in \spc space of the two best solutions in both runs.

\begin{table}[ht]
\centering
\begin{tabular}{p{5cm} r r r r }
Measurement & ud400 & grid400 & rand400 & normrand400 \\
\hline
Number of datasets with a single best selection & 45 & 59 & 16 & 12\\
Proportion of datasets with the same best solution for both runs &9\%& 
54\% & 6\% & 8\%\\
Average rank of the best solution in one run, in the other run & 20.2  &
 1.92  & 29.7 & 24.4\\
Median difference between accuracy of the best and second best
  selections in a single run  & 0.0009 & 0.0048 & 0.0013 & 0.0010\\
Median difference between the accuracy of the best solutions in both
  runs & 0.0089 &0.0085 & 0.012 & 0.011\\
Average distance between best selection in both runs in \spc space &
 4.7 & 1.2 & 3.9 & 3.9\\
 \hline
\end{tabular}
\caption{Comparison of the selection of the best hyperparameters for
  two runs of the \alg{ud400}, \alg{grid400}, \alg{rand400}, and \alg{normrand400} algorithms. See
  text for the explanation of the measurements.} \label{tworuns}
\end{table}

Let us inspect the result for the \alg{ud400} search algorithm. 45 datasets resulted in at least one run selecting a single best set of hyperparameters. Let us call $C_1$ and $\gamma_1$ the best selection for the first run and $C_2$ and $\gamma_2$ the best selection for the second run. $\est^1_1$ and $\est^1_2$ are the corresponding accuracies of the best (superscript 1) points for each run (subscript 1 and 2).  $\est^2_1$ and $\est^2_2$ are the accuracies of the second-best (superscript 2) solution for each run. For 9\% of the datasets, both runs resulted in the same solution, that is $C_1=C_2$ and $\gamma_1 = \gamma_2$.  The average rank of the accuracy at the point $C_2, \gamma_2$ in the first run and the point $C_1, \gamma_1$ in the second run was 20.2. That is, on average, the best point in one run ranks as the 20th best point in the other run. This seems the strongest evidence that the \alg{ud400} is mostly searching for peaks in the ``noise'' of the $\est_G$ surface - if the peaks in the accuracy were really peaks in the $\trs_{G}(C,\gamma)$ surface one would not expect such a large difference from one run to the other. This evidence can also be seen considering the accuracy values. The median difference between $\est^1_1$ and $\est^2_1$ (and $\est^1_2$ and $\est^2_2$), that is, the best and second best solutions of a single run is $8.6 \times 10^{-4}$. The median difference between $\est^1_1$ and $\est^1_2$, that is, the accuracy of the best solutions across runs, is ten times as large $8.9 \times 10^{-3}$. Again, one would not expect such a difference between the two runs. Finally, the average distance between $C_1, \gamma_1$ and $C_2, \gamma_2$ is 4.7.

The numbers are similar for the \alg{rand400} and \alg{normrand400} algorithms, but for them, there were fewer datasets that generated a single best solution. The exception is the \alg{grid400} algorithm, which has a much higher rate of agreement of the best solution between the two runs, and therefore a much lower difference between the median average between first and second-best in a single run, the difference of the best accuracy between runs, and the average distance between the best solutions between the two runs. We do not know how to explain this discrepancy of the grid search in relation to the others.

\subsection{Results on the different post search selection procedures}
\label{select-results}

There were 995 combinations of grid-like algorithms (\alg{grid}, \alg{ud}, \alg{rand}, \alg{normrand}) and subsets (half of each dataset) where there was more than one best solution for the hyperparameters. For these combinations, the median number of solutions was 5, the mean number of solutions was 24.57, the minimum 2, and the maximum 293.

Table~\ref{tab:select1} displays the mean rank of the accuracy on future data for each selection procedure. As one can see, there seems to be very little difference between the ranks. In fact, the Friedman test on the accuracy on future data of the selection procedures results in a p-value of 0.42 (chi-squared = 4.948, df = 5) which shows that there are no significant differences among the selection procedures.

\begin{table}[ht]
\centering
\begin{tabular}{lr}
  \hline
 selection procedure & mean rank \\ 
  \hline
  mingC & 1.40 \\ 
  maxCg & 1.41 \\ 
  randomCg & 1.41 \\ 
  meanCg & 1.41 \\ 
  minCg & 1.42 \\ 
  maxgC & 1.46 \\ 
   \hline
\end{tabular}
\caption{The mean rank in terms of accuracy on future data on the post search selection procedures}\label{tab:select1}
\end{table}

  It could be the case that the non-significant differences are due to a large number of comparisons. Table~\ref{tab:select2} only compares the ``more reasonable'' selection procedures: \alg{minCg} \alg{mingC}, \alg{randomCg}, and \alg{meanCg}. The resulting Friedman test has a p-value of 0.80 ( chi-squared = 1.0188, df = 3). Therefore, there is no evidence of differences among the post-search selection procedures.

\begin{table}[ht]
\centering
\begin{tabular}{lr}
  \hline
selection procedure & mean rank \\ 
  \hline
 mingC & 1.35 \\ 
 randomCg & 1.36 \\ 
 meanCg & 1.36 \\ 
 minCg & 1.38 \\ 
   \hline
\end{tabular}
\caption{The mean rank in terms of accuracy on future data of the four more reasonable post-search selection procedures}\label{tab:select2}
\end{table}

Finally, even if we restrict the number of solutions to 6 or above (which results in 464 combinations of grid-like algorithms and subsets) the p-value of the Friedman test for all 6 selection procedures is 0.63 (chi-squared = 3.4738, df = 5) and for the 4 selection procedures 0.56 (chi-squared = 2.0767, df = 3).

So, there is ample evidence that despite reasonable arguments to use one or another post-search selection procedure, there are no significant differences in future data accuracy among them.

\subsection{Hyperparameters selected by the best algorithms}
\label{sec:select-hyperp-best}

\begin{figure}[ht]
\begin{center}
\includegraphics[width=\textwidth]{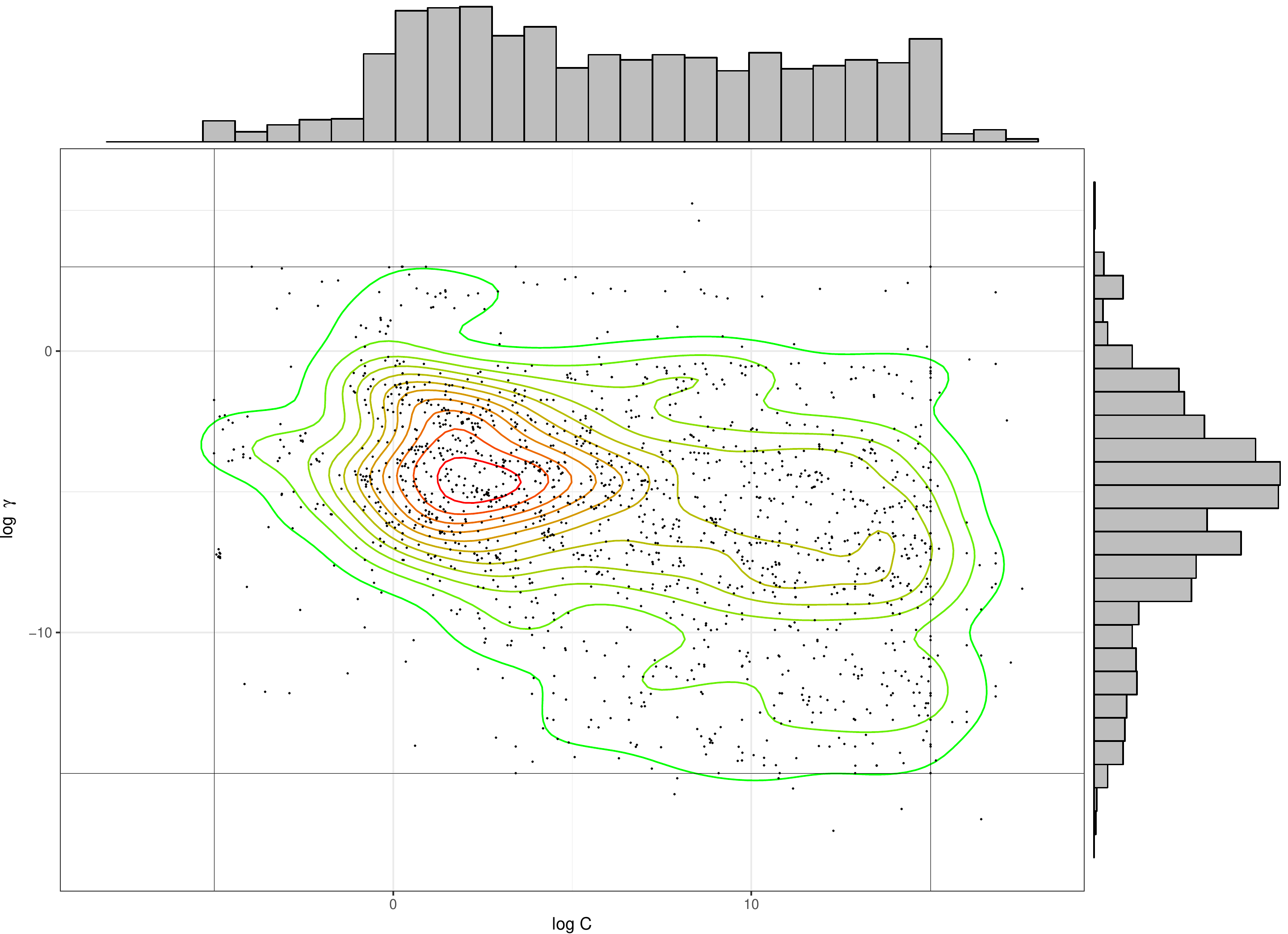}
\caption{The density plot of the $C$ and $\gamma$ selected by the best
  algorithms. The box are the constrained values }
\label{den1}
\end{center}
\end{figure}

Figure~\ref{den1} plots the best $\log C$ and $\log \gamma$ found by all the algorithms with 200 and 400 probes. The mean $\log C$ for the best algorithms without the \alg{normrand400} is 5.37 and the standard deviation is 5.33. The mean of the $\log \gamma$ is -5.56 and the standard deviation is 3.57.  These numbers are the source for our suggestion of the \alg{normrand} algorithm.  A mean of 5 and a standard deviation of 5 for the $\log C$ and a mean -5 and a standard deviation of 5 for the $\log \gamma$ are mnemonic approximations of those values.

\section{Discussion}
\label{sec:discussion}

An important issue to discuss is whether the results of this research can be generalized to any future dataset. The data sets used in this research are a large sample of those available at UCI in 2014, collected by the authors of \cite{delgado14}. We believe they can be considered as an unbiased sample of real-world data sets. But some classes of data sets are not represented in this set: very large data sets and data sets with a high number of dimensions. Fortunately, RBF SVM are usually not applied in these types of data sets: SVM is at least quadratic on the number of training examples and therefore it does not scale well to large data sets, with the possible exception of online SVM solvers; SVM with RBF kernel is usually not used on large dimensionality problems since for these cases the linear SVM is probably a better solution. Finally, all datasets are binary classification problems. We do not believe that for multi-class problems, where the SVMs are used in one-vs-one or one-vs-all configurations the conclusions would be different.

The choice of search algorithms was necessarily ad-hoc. We believe that this research covered most of the important proposal for SVM hyperparameter search algorithms. A somewhat non-conventional inclusion was the uniform design (\alg{ud}) search -- we wanted to include at least one algorithm from the research area of Design of Computer Experiments, which has not been very well represented in the hyperparameter search research.  The most notable absentee in this research are the many versions of sequential testing \citep{krueger2015fast}.

Also, we must point out that we made no attempt to select hyperparameters for the search procedures themselves. The more complex general optimization algorithms such as \alg{pso}, \alg{sa}, \alg{cma} have themselves hyperparameters: one has to define a function that will reduce the temperature in \alg{sa}; one has to define the number of particles and the relative strength of the momentum and the attractive force among the particles in \alg{pso}; and so on. Although we cited research that proposed using general optimization algorithms such as \alg{pso}, \alg{sa}, \alg{cma} and so on, \citep{pso1,pso2,sa1,sa2,sa3} we did not attempt to reproduce the algorithms hyperparameters as proposed by the authors or perform any search for those hyperparameters - we relied on the default of the implementations used. As we will discuss below, we do not think that a better choice of search hyperparameters would have resulted in a better selection of the SVM hyperparameters. Similarly, we do not believe that the inclusion of some of the missing search algorithms such as ant colony optimisation would have changed the main conclusions of this research in any significant way.

The time measurement results are less generalizable: the results reflect the 2019 implementation in R of the search algorithms, and the relative speeds of the different search algorithms may change as better and faster implementations become available. Also, we must point out that the timing data refer to the sequential execution of the search algorithms. Parallel execution is trivial for grid-like algorithms and there is published research on parallel versions of \alg{pso}, \alg{sa}, and \alg{bogp}.

We are not aware of any research published that systematically compares different search algorithms for SVM, with the exception of \cite{mantovani2015effectiveness} who compares on 70 data sets 6 different search algorithms: grid search, random search, default values for the hyperparameters, genetic algorithms (GA), estimation of distribution algorithms (EDA), and particle swarm optimization (pso). Different to this research, they run experiments of up to 10000 evaluations, but they use a non-standard cross-validation to tune and measure the accuracy of the SVM. What they call the S-CV (single cross-validation step) divides the data set into K subsets, where they use a 2-fold to train with the different combinations of hyperparameters, which are tested on one of the folds remaining. The last fold is only used once to measure the accuracy of the best selection of hyperparameters. The process is repeated K times, changing the last two folds and using the remained in training. We used the more traditional nested cross-validation - the subsets are an external 2-fold CV (to compute the accuracy) and the selection of the hyperparameters is performed by the internal 5-fold CV (one selection per fold). We do not know if the simpler, less costly S-CV, introduces bias to the evaluation, but we do know that nested CV is considered a more standard way of evaluating many classifiers \citep{cawley2010over}.  They performed the standard significance tests on the 6 search algorithms and did not find any significant differences except for the default settings which performed worse than the other 5 algorithms.

There has been some recent research on hyperparameter optimization in general \citep{hutter2011sequential,bergstra2011algorithms,bo2,bergstra2012random,krueger2015fast,automl} but they tend to focus on the more challenging problem of searching high dimensional spaces \citep{bergstra2013making}.

Finally, this research has some strong similarities to \citep{wainer2017empirical}. There the authors fixed the search procedure to select the SVM hyperparameters (an 11x10 grid search) and varied the resampling procedure used to compute the $\est_G$. That paper finds a similar conclusion that it is not worth it to use very precise and costly resampling procedures for the selection of hyperparameters. In particular, the authors suggest using a 2 or 3-fold resampling to select the hyperparameters. Here, we find that it is not worth it to use costly search procedures. In both cases, it seems that efforts in finding precisely the maxima in $\est_G$ or that measuring precisely $\est_G$ are wasteful.  The $\trs_G$ surface does not seem to have narrow, high maxima - narrow so that it is necessary to not only measure it precisely but also to spend a lot of search effort to find it, and high because the difference in accuracy between this high peak and the other not so good points is significant enough.  $\trs_G$ for SVMs seem to be smooth (at least near the maxima) and one should not expect that either measuring it more precisely or probing it many times will yield a ``sweet spot'' in accuracy. We can only speculate whether this might be true for other classification algorithms.

\section{Conclusions}
\label{sec:conclusion}

Regarding the main goal of this research, the comparison of different search algorithms for the $C$ and $\gamma$ hyperparameters of an RBF SVM, we found that:
\begin{itemize}
\item more evaluations of the cross-validated accuracy will, in general, result in the selection of hyperparameters that have higher cross-validated accuracy, which is not surprising.

\item if one is willing to allow many evaluations (100 or 400),  tree of Parzen estimators  \citep{bergstra2013making} and particle swarm optimization \citep{eberhart1995new} seem to be the two algorithms that achieve a good balance of execution time and finding higher accuracy points in the $\est_G$ surface.
\item Bayesian optimization \citep{bo1,bo2} does find high accuracy points but at a very high computational cost.

\item all grid-like algorithms \alg{grid}, \alg{rand}, \alg{ud}, and \alg{normrand} seem to have similar performance, but the accuracy gains for \alg{normrand} seems a little higher than the other three.

\item the hierarchical versions of \alg{grid} and \alg{ud} are, not surprisingly, good alternatives to the flat versions of the algorithms.

\item at lower evaluation points (25) \alg{nelder} and \alg{pso} seem to be the best alternative

\item at even lower number of evaluations (10 and 20) the default value of the $\gamma$ parameter selected by \alg{sigest}  \citep{caputo2002appearance} seems a better solution than the alternatives proposed by Sklearn and distance between two classes. The \alg{sigest} default value for $1/\gamma$ tested in this research is the median distance squared among a 50\% sample of the data points.

  \item the SVM-specific algorithms \alg{svmpath} \citep{path1} and \alg{asym} \citep{asymptotic} are not competitive alternatives.
  
  \end{itemize}

 Regarding the second goal of this research, efforts in finding better hyperparameters with higher values of the cross-validated accuracy seems not to be correlated to better expected accuracy on future data. There is no important difference in (an estimative of) the expected accuracy on future data among the best performing algorithm and the \alg{grid100}. 

Regarding the third goal of this research, we could not find any significant difference between the different post-search selection procedures. That is, despite arguments that one should select a lower $C$ and lower $\gamma$ if multiple best solutions were found, there is no evidence that this selection procedure is any better than a random selection, or other different procedures in terms of accuracy on future data. 

Regarding the fourth goal of this research, at a first approximation, the distribution of the best $\log C$ (for datasets whose features are scaled to zero mean and unit standard deviation) is a normal distribution of mean equal to 5 and standard deviation equal to 5. And  the best $\log \gamma$ follows a  normal distribution with mean equal to $-5$ and standard deviation of 5.

Finally, 
an important conclusion of this paper is that is in likely not worth it to spend too much computational time searching for better hyperparameters. It is likely that the search algorithms are finding ``noisy'' peaks in the $\est_G$ surface which are not representative of ``real'' peaks in the $\trs_G$ surface. If that is indeed the case, then one should not spend a lot of computational effort searching for the best set of hyperparameters, and our conclusions that \alg{pso} and \alg{tse} are, in general, the best search algorithms for a high number of evaluations are only of theoretical interest. In this case, one should choose algorithms with a lower number of evaluations. Given that grid like algorithms can be trivially parallelized, if the user has enough computational instances, we believe that a grid-like search with 25 or a few more evaluations is likely enough to find good hyperparameters for most of the problems.



\end{document}